\documentclass[letterpaper, 10 pt, conference]{ieeeconf}      
                   
\IEEEoverridecommandlockouts  
\usepackage{cite}
\usepackage[font=footnotesize]{subfig}
\usepackage{multirow}
\usepackage{comment}
\usepackage{ifpdf}
\usepackage[pdftex]{graphicx}
\usepackage{epstopdf}
\usepackage[cmex10]{amsmath}
\usepackage{algorithmic}
\usepackage{array}
\usepackage{mdwmath}
\usepackage{mdwtab}
\usepackage{eqparbox}
\usepackage{verbatim}
 \usepackage[ruled,vlined]{algorithm2e}
\usepackage[font=footnotesize]{subfig}
\usepackage{fixltx2e}
\usepackage{stfloats}
\usepackage{comment}
\usepackage{amsmath,bm}
\usepackage{amssymb}
\usepackage{color}
\usepackage{subfig}
 \usepackage{makecell} 
\usepackage{multirow}
\usepackage{float}
\usepackage{caption}
\usepackage{comment}
\usepackage{hyperref}
\begin{document}

\title{ \LARGE \bf Hierarchical Semi-Supervised Learning Framework for Surgical Gesture Segmentation and Recognition Based on Multi-Modality Data
}
\author{Zhili Yuan, Jialin Lin, Dandan Zhang
\thanks{All authors are with the Department of Engineering Mathematics, University of Bristol, Bristol, United Kingdom. }}

\maketitle

\begin{abstract}
Segmenting and recognizing surgical operation trajectories into distinct, meaningful gestures is a critical preliminary step in surgical workflow analysis for robot-assisted surgery. 
This step is necessary for facilitating learning from demonstrations for autonomous robotic surgery, evaluating surgical skills, and so on. In this work, we develop a hierarchical semi-supervised learning framework for surgical gesture segmentation using multi-modality data (i.e. kinematics and vision data). More specifically, surgical tasks are initially segmented based on distance characteristics-based profiles and variance characteristics-based profiles constructed using kinematics data.  Subsequently, a Transformer-based network with a pre-trained `ResNet-18' backbone is used to extract visual features from the surgical operation videos. 
By combining the potential segmentation points obtained from both modalities, we can determine the final segmentation points. Furthermore, gesture recognition can be implemented based on supervised learning.

The proposed approach has been evaluated using data from the publicly available JIGSAWS database, including Suturing, Needle Passing, and Knot Tying tasks. The results reveal an average F1 score of 0.623 for segmentation and an accuracy of 0.856 for recognition. For more details about this paper, please visit our website: \url{https://sites.google.com/view/surseg/home}. 
\end{abstract}



\section{Introduction}
The rapid development of machine learning leads to substantial growth in the field of surgical data science and robot-assisted surgery \cite{zhang2022teleoperation,wang2021real, maier2022surgical, zhang2018self}. Surgical tasks mainly rely on the repetition and execution of specific gestures, while they can be further decomposed into basic surgical gestures. 
The segmentation and recognition of surgical gestures is an important task for surgical workflow analysis, which can benefit the training and assessment of surgeons \cite{zhang2020automatic}, enable semi-autonomous robotic surgery via learning from demonstration, and provide real-time feedback and guidance for surgeons to enhance the efficiency of robotic surgery \cite{chen2020supervised}. 

\begin{figure}[htpb]
\centering
\captionsetup{font=footnotesize,labelsep=period}
  \includegraphics[width = 1\linewidth]{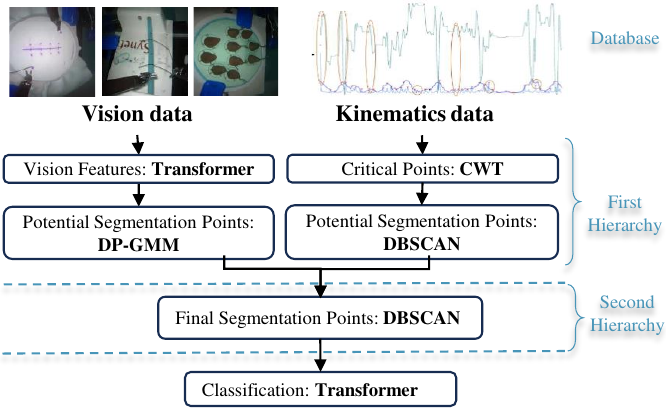}
  \caption{Flowchart of the proposed hierarchical semi-supervised learning framework for surgical gesture segmentation and recognition. }
  \vspace{-0.6cm}
  \label{flowchart}
\end{figure}

Supervised learning has been widely used for surgical gesture segmentation and recognition  \cite{van2021gesture}. Traditional machine learning algorithms, such as Hidden Markov Models (HMM) \cite{bhuyan2014novel}, Linear Dynamical Systems (LSD), Conditional Random Fields (CRF), and Markov/semi-Markov CRF (MsM-CRF), have been used for surgical gesture segmentation and recognition \cite{tao2013surgical}.
Recurrent Neural Networks (RNNs) and their variations \cite{zhang2021surgical}, such as Long-Short Term Memory (LSTM), and Gated Recurrent Units (GRU),  have also been widely used for gesture segmentation, since they have been proven to be effective for processing sequential data \cite{dipietro2019segmenting}. 
However, the machine learning-based gesture segmentation quality significantly depends on the availability of large-scale labeled datasets.

Unsupervised learning has been explored for surgical gesture segmentation. For example,  Transition State Clustering (TSC) has been applied to the surgical gesture segmentation task \cite{krishnan2017transition}. Gaussian Mixture Model has been used to segment surgical gestures based on kinematic data and the visual features extracted by pre-trained Convolutional Neural Networks (CNN) \cite{murali2016tsc}. A Dense Convolutional Encoder-Decoder Network (DCED-Net) has been combined with Temporal Convolutional Networks (TCN) to further enhance the segmentation accuracy \cite{zhao2018fast}. However, the performance of unsupervised learning methods is not desirable compared to supervised learning methods.
As a compromise between data labeling workload and model performance, we investigate semi-supervised learning in this paper.

Semi-supervised learning has been applied to the segmentation of surgical gestures using kinematic data. For instance, within a semi-supervised learning framework, the Stacked Denoising Autoencoder (SDAE) has been utilized for feature extraction in an unsupervised manner \cite{gao2016unsupervised}, while Dynamic Time Warping (DTW) was employed to align the kinematic data from different trials of the surgical task. Subsequently, a voting mechanism based on kernel density estimation was applied to transfer labels from template trials to the test trial, resulting in gesture recognition through semi-supervised learning. An RNN-based generative model has been developed for surgical gesture recognition, using  only one annotated sequence \cite{dipietro2019automated}. Its performance outperforms other approaches, including the RNN-Based Autoencoder and RNN-Based Future Prediction. However, most of the methods mentioned above only use kinematic data to implement gesture segmentation or recognition. Recent literature has demonstrated 
that the performance of surgical gesture recognition can be improved using multi-modality data compared with their single-modality counterparts \cite{murali2016tsc}.  

To this end, we propose a hierarchical semi-supervised learning framework for surgical gesture segmentation and recognition using multi-modality data, which aims to eliminate the need of collecting a large amount of labeled data for supervised learning while ensuring high segmentation and recognition performance. The main contributions are listed as follows.
\begin{itemize}
    \item We propose a hierarchical semi-supervised learning framework for surgical gesture segmentation, utilizing both kinematics and video data. Our proposed method was compared to state-of-the-art approaches, with  results indicating higher accuracy of segmentation.
    \item We employ a transfer learning approach to extract useful features from a limited amount of labeled surgical video data, thus ensuring high data efficiency in our proposed method. This method utilizes a Transformer-based architecture, with ResNet-18 serving as the backbone.    
\end{itemize}

\section{Methodology}

\subsection{Overview}



The core target of surgical gesture segmentation is to locate the starting point and ending point of a specific surgical gesture \cite{chakraborty2018review}.
Surgical gestures involve dynamic movements that feature transitions encompassing both local and global motions. As such, kinematic data play a crucial role in detecting transitional states and is thus considered valuable for surgical gesture segmentation. 
Simultaneously, vision data, with  contextual information, contributes to improving the accuracy of surgical activity segmentation and recognition. Therefore,  multi-modality data will be used to enhance surgical gesture segmentation and recognition accuracy in this paper. 

We define features initially identified from specific feature extraction methods as \textbf{`critical points'} (also known as `change points' or `transition points' in other papers). It is worth noting that the occurrence of critical points may not be simultaneous for different characteristics profiles.  For instance, in a Knot Tying task, orienting a needle may only lead to changes in the rotation, while the translation  may remain unchanged. 
As a result, we introduce a hierarchical structure to address this issue.  In the first layer, multiple meaningful features extracted from kinematic data are clustered using `Density-Based Spatial Clustering of Applications with Noise (DBSCAN)', while all visual features extracted by the Transformer-based model are clustered with  Dirichlet Process Gaussian Mixture Model (DP-GMM). 
The resulting data will be referred to as \textbf{`potential segmentation points'} (also known as `pre-segmentation points' in other papers) throughout this paper. In the final step, the second layer of DBSCAN is applied to all potential segmentation points to generate \textbf{`final segmentation points'}.  \textbf{Algorithm \ref{overview}} describes the whole segmentation method.

\begin{algorithm}
\renewcommand{\algorithmicrequire}{\textbf{Input: Raw kinematics and vision data}}
\renewcommand{\algorithmicensure}{\textbf{Output: Estimated segmentation points}}
\begin{algorithmic}[1]
\FOR{All kinematics data}
\STATE Obtain Filtered Kinematics Data: \\ Raw Data $\leftarrow$ Kalman and Savitzky-Golay filters
\STATE Obtain critical points $cp^{ori}$: Distance Characteristics-Based Profiles $\leftarrow$ CWT
\STATE Obtain critical points $cp^{var}$: Variance Characteristics-Based Profiles $\leftarrow$ CWT \\
(see Section \ref{profiles})
\ENDFOR

\FOR {All vision data}
\STATE Extract visual features $f_{vis}$:  \\Preprocessed video data $\leftarrow$  Transformer \\
(see Section \ref{pre-vision})
\STATE Obtain all potential segmentation points for vision data $CP_{vis}$:\\  $f_{vis}$ $\leftarrow$ DP-GMM
(see Section \ref{pre-vision-GMM})
\ENDFOR
\STATE Obtain all potential segmentation points for kinematics data $CP_{kine}$: \\ $[cp^{ori}$, $cp^{var}]$ $\leftarrow$ DBSCAN
$1_{st}$ Layer \\
\STATE Obtain final segmentation points $esp$: \\ $[CP_{kine}$, $CP_{vis}]$ $\leftarrow$ DBSCAN $2_{nd}$ Layer \\
(See Section \ref{final})
\caption{Overview of the Proposed Framework}
\label{overview} 
\end{algorithmic}
\end{algorithm}

One of the well-known surgical activity datasets is the JHU-ISI Gesture and Skill Assessment Working Set (JIGSAWS) \cite{gao2014jhu}, where a large number of  surgical operation data was collected from surgeons using the da Vinci Surgical Robot.  This dataset contains kinematics and video data along with manual annotation with gestures' name and operators' skill levels. There are three surgical tasks in the dataset, including 39 suturing,
36 knot tying and 20 needle passing demonstrations.  The definitions of the surgical gesture for all three tasks can be found in \cite{gao2014jhu}. The JIGSAWS dataset is used for model training and evaluation in this paper.

\subsection{\textbf{Pre-Segmentation Based on Kinematics Data}}
\label{profiles}

\subsubsection{\textbf{Characteristics Profile }Construction}~{}\\
The kinematic characteristics of the surgical gesture transition points can differ from other points within the same gesture \cite{guo2017rrv}. That is to say, kinematic data points are clustered when they are transitioning to another gesture, while smooth intervals represent an in-motion state.
By calculating the translation and rotation distance between two consecutive points along the trajectory, we can find out critical points (transition points) in the density of trajectories. In addition, a variance characteristics-based method is adopted to enable the segmentation to be more robust to noise \cite{tsai2019unsupervised}.
The raw kinematics data recorded include noises. Therefore, the Kalman filter is applied to the raw trajectories for noise removal. 
Following that, new profiles can be constructed from distance and variance characteristics-based profiles. The Savitzky–Golay filter can then be applied to further smooth the newly constructed profiles after normalization.

 Denote the original frame for the kinematics data as ${B}$, Translation Matrix as $\mathbf{T(t)}$, and 3$\times$3 Rotation Matrix  as $\mathbf{R}(t)$. Suppose that $\mathbf{Q(t)}$ is the quaternion converted from $\mathbf{R}(t)$, which also represents  the end-effector's orientation in the original dataset. The corresponding Euler angles converted by $\mathbf{R}(t)$ is $\mathbf{e}(t)=[e_x(t),e_y(t),e_z(t)]$. Let $\mathbf{p}(t)=[p_x(t),p_y(t),p_z(t)]$ denote the end-effector position at time step $t$,   $\mathbf{\dot{p}}(t) = [\dot{p_x}(t), \dot{p_y}(t), \dot{p_z}(t)]$ denote the end-effector velocity  at time step $t$.  

\textbf{Distance Characteristics-Based Profiles:}
Translation and rotation distance between consequent time steps can be calculated for both left and right surgical tools, respectively. 
For the translation component, the Euclidean distance of the end-effector between time step $t$ and $t-1$ can be calculated by $ D_{trans} = \vert\vert{\mathbf{p}(t)-\mathbf{p}(t-1)}\vert\vert$. As for the rotational component, the distance is determined by $D_{rot} = arccos(2   (\mathbf{Q}(t)\cdot\mathbf{Q}(t-1)) ^ 2 - 1)$.

Four distance characteristics-based profiles can be constructed, including the i) left-hand translation distance profile, ii) left-hand rotation distance profile, iii) right-hand translation distance profile, and iv) right-hand rotation distance profile. These newly constructed profiles are normalized before being used to identify critical points.

\textbf{Variance Characteristics-Based Profiles:}

Relying solely on translation and rotation distances may lead to over-segmentation. To address this issue, the variance characteristics-based approach can be used to capture the essential features of kinematic data \cite{tsai2019unsupervised}.  Segmentation points tend to cluster, while points within continuous motion are typically sparse \cite{tsai2019unsupervised}. These characteristics are frame invariant, indicating that they remain consistent when the data is transformed into a new frame of reference. 
Hence, calculating variances across all new frames can help identify significant changes. High variance points suggest significant changes in terms of angle or angular speed values. Understanding these variance characteristics can help differentiate between continuous motion and transition periods, thus enhancing the segmentation accuracy.

Assume there are \( N \) new frames generated from random translation matrix \(\mathbf{^{n}T}(t)\) and rotation matrix \(\mathbf{^{n}R(t)}\) separately for \( n=1,2,\ldots,N \). The trajectory points in the new frames could be written as $\mathbf{^{n}p}(t) = \mathbf{^{n}R(t)} \mathbf{p}(t) + \mathbf{^{n}T}(t)$.
Given the rotation matrix \(\mathbf{^{n}R(t)}\), the Euler angles \(\mathbf{e}(t)\) in the new rotation frame for each \( n \) can be derived. The variance characteristics-based profile can be calculated based on all new frames at each time step, as illustrated in  \eqref{varequ}:
 \begin{equation}
 \begin{split}
Var^n_{trans}(t) = Var[^{n}p_x(t)]+Var[^{n}p_y(t)]+Var[^{n}p_z(t)]\\
 Var^n_{rot}(t) = Var[^{n}e_x(t)]+Var[^{n}e_y(t)]+Var[^{n}e_z(t)] 
 \label{varequ}
 \end{split}
 \end{equation}

Four variance characteristics-based profiles can be constructed, including the i) left-hand translation variance profile, ii) left-hand rotation variance profile, iii) right-hand translation variance profile, iv) right-hand rotation variance profile.


\subsubsection{\textbf{Critical Points} Determination}
Since segmentation points separate two gestures, they typically occur when there is a significant distance between two consecutive points along a trajectory.  
That is to say, the critical points normally appear at the corners of the peaks among the newly constructed characteristics profiles. 
They can thus be identified
when two points have a large distance among all the profiles. 
All the critical points obtained from different profiles can be combined as a point set, which can be later used for potential segmentation points identification.

Two methods can be used to identify these peaks and corners for determining the critical points. The first method involves specifying a threshold height and prominence, while the second method involves using continuous wavelet transformation (CWT) to identify relative peaks and corners. Both methods have been widely used in the literature for identifying critical points in various types of data.  The comparisons of these two methods can be found in the supplementary materials on our website.

\begin{figure*}[htbp]
  \centering
  \captionsetup{font=footnotesize,labelsep=period}
  \includegraphics[width = 0.95\textwidth]{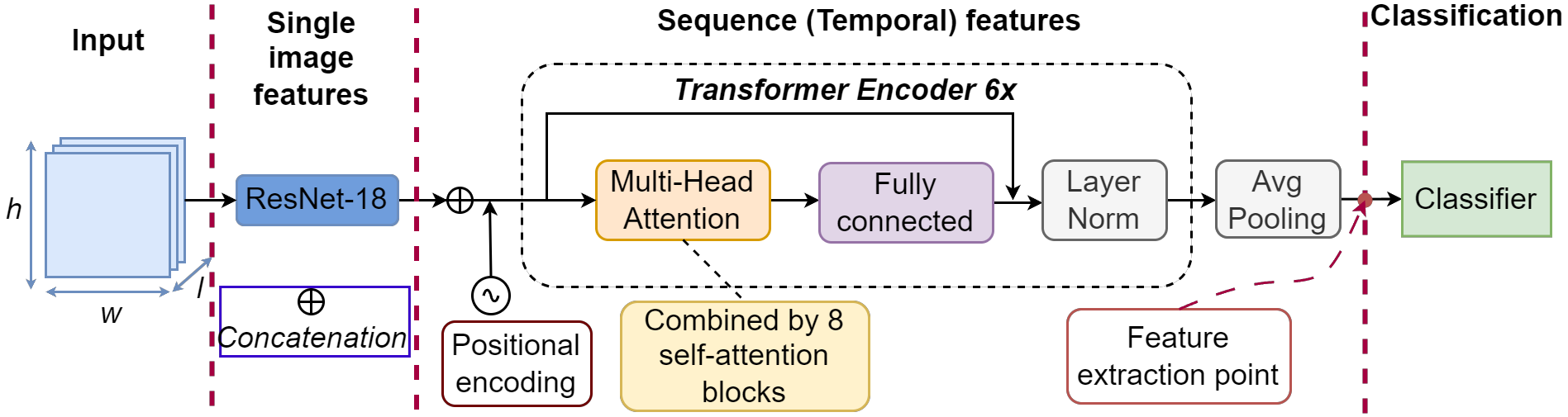}
  \caption{Network Architecture of the Transformer-based network with a pre-trained `ResNet-18' backbone for visual features extraction.} 
   \vspace{-0.4cm}
  \label{transnet}
\end{figure*}

\subsection{\textbf{Pre-Segmentation Based on Vision Data}}
\subsubsection{Transformer-Based Feature Extraction}
\label{pre-vision}
Deep neural networks have demonstrated their immense capabilities in the field of computer vision. More recently, Transformers have been proven to outperform CNNs and RNNs when they are applied to sequential data processing, since they can deal with long-range context dependencies \cite{d2020transformer}. Transformers have the capacity to establish temporal links between the present and past frames based on self-attention mechanism \cite{czempiel2021opera}. Therefore, a modified Transformer-based architecture (see Fig. \ref{transnet}) is integrated with one of the Deep Residual Neural Network model-based backbones to implement visual feature extraction in this paper. Specifically, the ResNet-18 architecture with pre-trained weights from ImageNet is used as the backbone for the model, which accelerates the model training process. The remaining weights in the architecture are fine-tuned using the limited labeled data in JIGSAWS \cite{d2020transformer}. 





To prepare for model training, the video data is initially transformed into a sequence of images. For training the Transformer-based model, every input consists of a series of $l$ images that share the same gesture label. The value of $l$ can be referred to as the sliding window size, representing the length of a sequence of images used as model input. All inputs have $(l-s)$ overlapping frames, where $s$ represents the step size for sequential data generation. In this paper, $l=30$ and $s=5$ are used to construct a new dataset with a reorganized data structure for model fine-tuning. After extracting the features from the vision data, we obtain a new dataset that includes sequences of labeled data represented by feature vectors.

\subsubsection{Feature Clustering Based on Unsupervised Learning}
\label{pre-vision-GMM}

The feature clustering method used in this work is inspired by \cite{krishnan2017transition}, where an unsupervised Transition State Clustering (TSC) algorithm was used to identify potential segmentation points between two Gaussian clusters. It has been demonstrated that DP-GMM, learning through Expectation-Maximization, exhibits excellent performance in clustering high-dimensional data without prior knowledge of the ground truth cluster number.


If there is a significant difference between two adjacent features, it is likely that the gestures change at that point, and thus the potential segmentation points could be found. The visual critical points extraction is performed using two layers of DP-GMM.  The first layer clusters across all the frames to find as many critical points as possible, while the second layer further clusters those critical points to determine the potential segmentation points.




For the DP-GMM method, Dirichlet Process is a probability distribution that resolves the issue of requiring a pre-determined number of classes when implementing a pure GMM. Suppose there are $N$ data points $[x_1,x_2,...,x_N]$, where each is generated from a different distribution $g_i = g_1,g_2,...,g_N$, and each distribution has corresponding parameters $\theta_i = {\theta_1,\theta_2,...,\theta_N}$.  Assuming each $g_i$ follows a distinct Gaussian distribution, $\theta_i$ follows a certain continuous distribution $H(\theta)$.
The Dirichlet Process involves constructing a discrete distribution $G$ to make $\theta_i \sim G$, where $G \sim DP(\alpha,H)$. Here, $\alpha$ is known as the concentration parameter, which controls the shape of the distribution.

\subsection{\textbf{Final Segmentation Point Identification}}
\label{final}
Density-Based Spatial Clustering of Applications with Noise (DBSCAN) is a clustering algorithm that can be used to cluster points with sufficient density. Unlike other clustering algorithms, it does not require the number of clusters to be specified in advance and only requires two hyperparameters.
DBSCAN is also capable of identifying and handling abnormal values, which is effective in eliminating misidentified critical points or potential segmentation points.

In this paper, two layers of DBSCAN were utilized to locate segmentation points. The first layer is used to cluster all the kinematic critical points and therefore obtain kinematic potential segmentation points. The second layer combines the output of both kinematic and vision potential segmentation points to identify the final segmentation points. To implement this algorithm, two parameters must be defined: the minimum number of points ($minp$) required to determine a dense region, and the maximum distance ($eps$) between two points.
 Potential and final segmentation points should have a sufficient number of other data points (defined by $minp$) within a specified distance (defined by $eps$). Boundary points refer to data points that are not core points but lie within the neighborhood of a core point.
 If a point does not belong to either of these categories, it is classified as a noise point.

\section{Experiments and Results Analysis}

\begin{figure}[htpb]
\centering
\captionsetup{font=footnotesize,labelsep=period}
  \includegraphics[width =\linewidth]{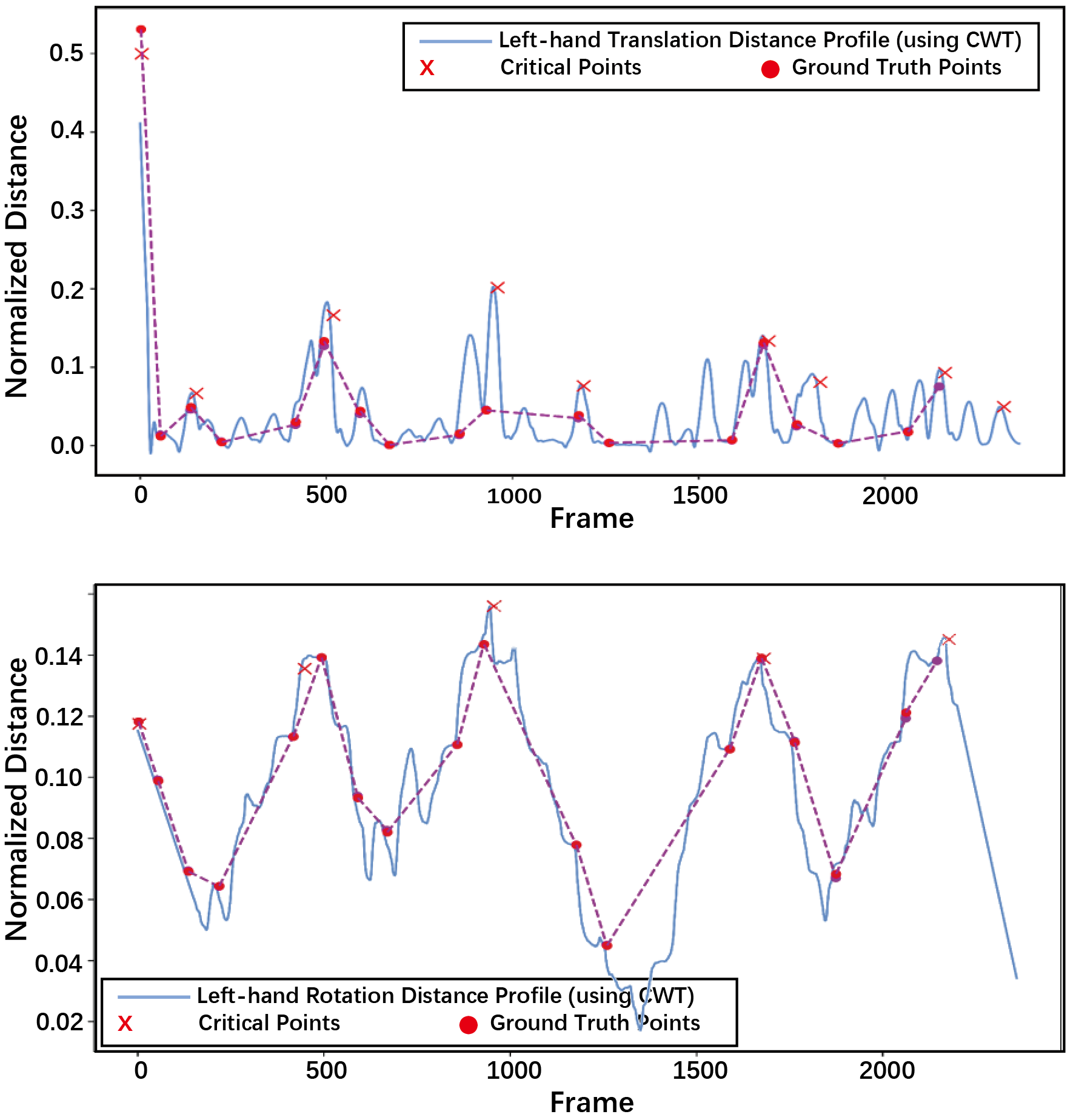}
  \caption{Critical points determined based on the left-hand translation distance profile and left-hand rotation distance profile. (Data Source: Suturing File F004 in JIGSAWS)}
  \label{left}
  \vspace{-0.6cm}
\end{figure}

\subsection{Experiment Design}
The dataset used in this study consists of a total of 103 demonstration files, which are divided into a training set, a validation set, and a test set in a 69:21:13 ratio. During the model training process for the Transformer network, images are cropped to 80x80 pixels to accelerate the network training process and accommodate CUDA memory. To prevent overfitting, we adopted data augmentation techniques such as random rescaling, random cropping, and random rotation within a range of -15 to 15 degrees \cite{d2020transformer}. The batch size is set to 30, and the total number of training epochs is 80. We use the Adam optimizer with weight decay, and the loss function is constructed using categorical cross-entropy. The initial learning rate is set to 0.001


Bayesian optimization is used to automatically find the optimized values of hyperparameters \cite{frazier2018tutorial}. Specifically, the concentration parameters for the two DP-GMM layers (denoted as $\alpha1$ and $\alpha2$), the component numbers ($n_1$ and $n_2$) in the two Dirichlet Processes, and the maximum distance $eps$ for DBSCAN can be fine-tuned through Bayesian optimization. The F1-score is chosen as the target function for optimization. The results of the optimized hyperparameters found through Bayesian Optimization are summarized in Table \ref{para}.

\begin{table}[htpb]
\centering
\caption{Optimized Hyperparameters Determined Based on  Bayesian Optimization}
 \begin{tabular}{ c |c | c | c | c |c  } 
 \hline
\textbf{Task} &  \textbf{$n_1$} & \textbf{$n_2$} & \textbf{$\alpha1$} & \textbf{$\alpha2$} & \textbf{$eps$}   \\ \hline
Suturing & 768 & 145 & 9.437 & 3.729  & 28\\ \hline
Needle Passing  & 545 & 708 & 183.5 & 4.489 & 15.49 \\ \hline
Knot Tying &  658 & 445 & 157.8 & 162.2 & 31.76 \\ \hline
\end{tabular}
\label{para}
\end{table}

Three standard frame-wise metrics (precision, recall, and F1-score) are used to evaluate the segmentation results \cite{van2021gesture}.
As for multi-class classification, the micro and the macro average are calculated based on the confusion matrix as shown in the following equations.
\begin{gather}
Micro_{precision} =\frac{\sum_{i=1}^{n}TP_i}{{\sum_{i=1}^{n}TP_i + \sum_{i=1}^{n}FP_i}} \\
Micro_{recall} =\frac{\sum_{i=1}^{n}TP_i}{{\sum_{i=1}^{n}TP_i + \sum_{i=1}^{n}FN_i}}\\
Macro_{precision} =\frac{1}{n}\sum_{i=1}^{n}P_i \\
Macro_{recall} =\frac{1}{n}\sum_{i=1}^{n}R_i
\label{class_metric}
\end{gather}
where   $TP_i$, $FP_i$,$FN_i$ denotes True Positive, False Positive, and False Negative for each class $i$,  $P_i$, $R_i$ denotes the precision and recall for each class $i$.

\subsection{Kinematics-Based Segmentation}

Smooth trajectory profiles can be obtained after applying the Kalman filter, as oscillation points are eliminated, and critical points can be identified with higher accuracy. We conducted follow-up experiments and discovered that critical points were more accurately identified after applying the Savitzky-Golay filter.

\begin{figure}[htpb]
  \centering
  \captionsetup{font=footnotesize,labelsep=period}
\includegraphics[width =1\linewidth]{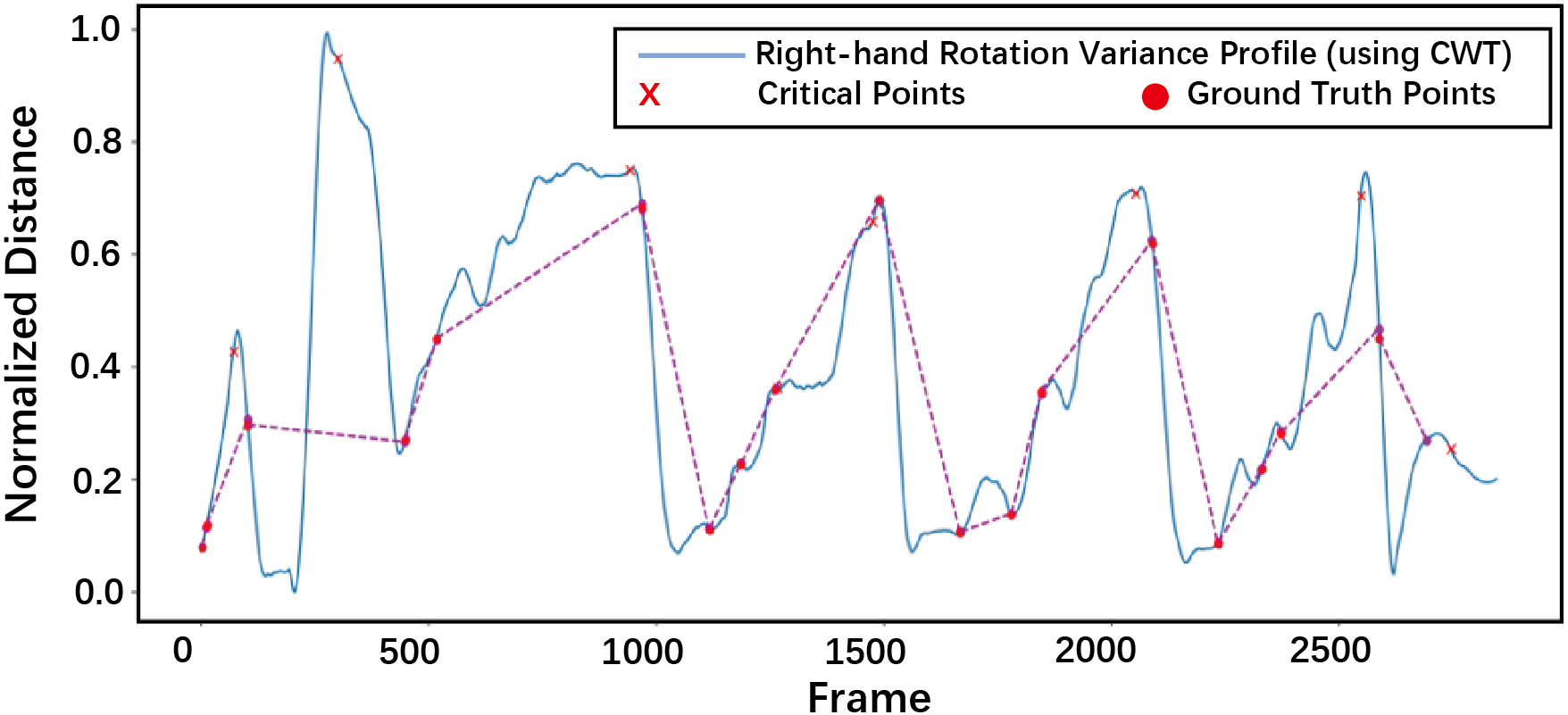}
  \caption{An example of right-hand rotation variance profile (Data Source: Suturing File B005, JIGSAWS)}
   \vspace{-0.4cm}
  \label{varright}
\end{figure}

Fig. \ref{left} provides an example of critical points extracted from the left-hand translation and rotation distance profiles using CWT. A comparison of the critical segmentation points obtained from the two profiles suggests that the rotation distance profile provides more accurate results than the translation distance profile. This may be due to the fact that the Suturing Task involves more rotational motion than translational motion, resulting in more prominent critical points in the rotation distance profile. Critical points identified from the rotation variance profile are shown in Fig. \ref{varright} as examples.

By combining all the \textbf{critical points} obtained from both distance characteristics-based and variance characteristics-based profiles, we can further obtain a set of \textbf{potential segmentation points} after applying the first layer of DBSCAN. Using these potential segmentation points, we assessed the kinematics-based segmentation with the following metrics:
i) mean recall, ii) precision, and iii) F1 score. For the Knot Tying task, these metrics are 0.482, 0.251, and 0.330, respectively. Similarly, for the Suturing Task, the mean recall, precision, and F1 score are 0.488, 0.242, and 0.324, respectively. For the Needle Passing task, the corresponding values are 0.372, 0.145, and 0.210, respectively.


The aforementioned results reveal that though the proposed method can correctly identify many segmentation points, there is still a noticeable deviation between the ground truth segmentation points and the potential segmentation points when relying solely on kinematics data.
One possible explanation is that the dataset is captured by surgeons with varying robotic surgical experience. Some surgeons' operation data may has a significant smoothness discrepancy in the trajectory. The smoother the trajectory is, the shorter pause it might contain between different gestures, and there might not be noteworthy critical points. Thus, the use of visual data is necessary to help improve segmentation accuracy, which contains some important context information to help differentiate different gestures.

\subsection{Vision-Based Segmentation Results}

\begin{figure*}[htpb]
  \centering
  \captionsetup{font=footnotesize,labelsep=period}
  \includegraphics[width=0.9\textwidth]{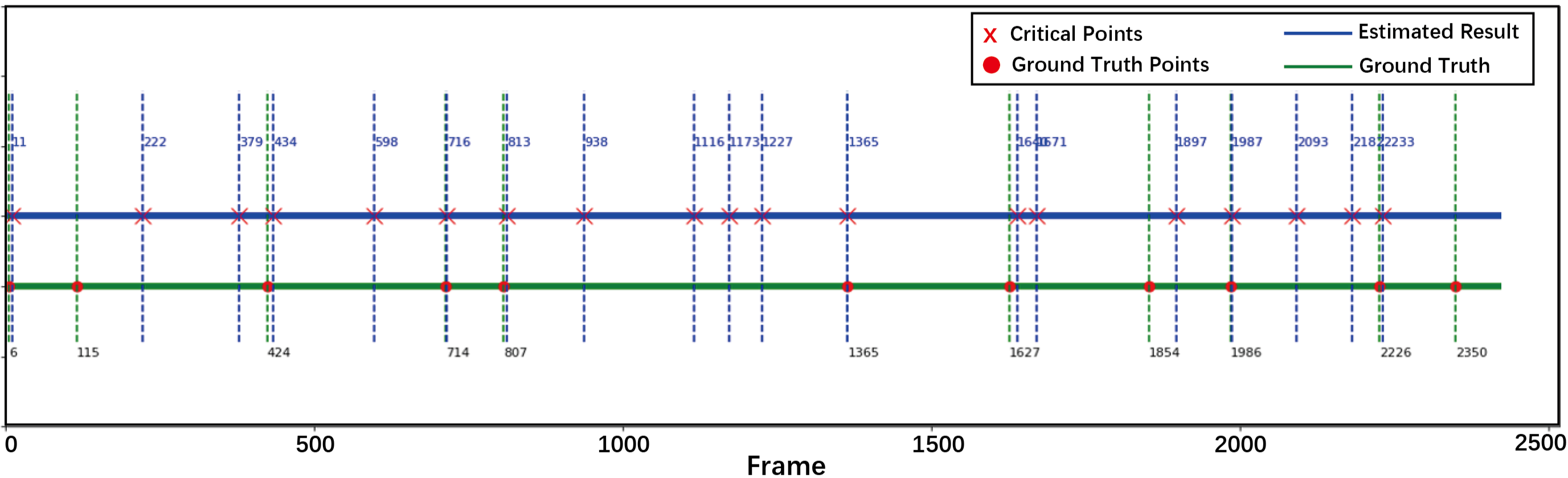}
  \caption{An example of the segmentation result using vision data (Data Source: Knot Tying File I002, JIGSAWS)}
   \vspace{-0.3cm}
  \label{vis_only}
\end{figure*}

 The testing accuracy of the Transformer-based model on the test dataset is 0.93. The mean of recall, precision, and F1-score for all Suturing segmentation metrics are 0.516, 0.634, and 0.550, respectively. As for Needle Passing and Knot Tying task, the scores for all the evaluation metrics are summarized in Table \ref{score_table}.  Take the Knot Tying task as an example, the segmentation results based on vision data are shown in Fig. \ref{vis_only}. 

 The accuracy of the segmentation points for the Needle Passing task was found to be the lowest among the three tasks. This could be attributed to the fact that most gestures for the Needle Passing task were conducted based on local operation, resulting in less significant differences in the distance characteristics-based profile compared to the other tasks. Additionally, critical features such as the needle pose and thread in the video data may be obscured, making the accurate identification of segmentation points challenging.

\subsection{Vision and Kinematics Fusion-Based  Results}

\begin{table}[t]
\centering
\caption{Summary of Segmentation Results}
 \begin{tabular}{c | c | c | c |c} 
 \hline
\renewcommand\arraystretch{0.2}
\textbf{Modality} & \textbf{Task} & \textbf{Recall} & \textbf{Precision} & \textbf{F1 score} \\ \hline
\multirow{3}{*}{ Kinematics} & Suturing & 0.488 & 0.242 & 0.324\\ \cline{2-5} 
& Needle Passing & 0.372 & 0.145 & 0.210\\ \cline{2-5} 
& Knot Tying & 0.482 & 0.251 & 0.330 \\ \cline{2-5}  \hline

\multirow{3}{*}{Vision} & Suturing & 0.516 & 0.634 & 0.559\\ \cline{2-5}
& Needle Passing & 0.415 & 0.620 & 0.488\\ \cline{2-5}
& Knot Tying & 0.567 & 0.673 & 0.615 \\ \cline{2-5}\hline
\multirow{4}{*}{\makecell{kinematics \\+ Vision }} & Suturing & 0.533 & 0.770 & 0.630 \\ \cline{2-5}
& Needle Passing & 0.510 & 0.647 & 0.570 \\ \cline{2-5}
& Knot Tying & 0.569 & 0.780 & 0.657 \\ \cline{2-5}
 & mean  & 0.537 & 0.745 & 0.623 \\ \cline{2-5} \hline
\end{tabular}
   \vspace{-0.3cm}
\label{score_table}
\end{table}

\subsubsection{Final Segmentation Results}
For the Needle Passing task, the average segmentation results in terms of recall, precision, and F1-score are 0.510, 0.647, and 0.570, respectively. For the Knot Tying task, the corresponding results are 0.569, 0.780, and 0.657, respectively. A comprehensive summary of all segmentation scores is presented in Table \ref{score_table}. These findings illustrate the superior performance of our proposed method in surgical gesture segmentation, as our framework achieves improvements in both recall and precision.

\begin{figure*}[htpb]
  \centering
  \captionsetup{font=footnotesize,labelsep=period}
  \includegraphics[width=0.9\textwidth]{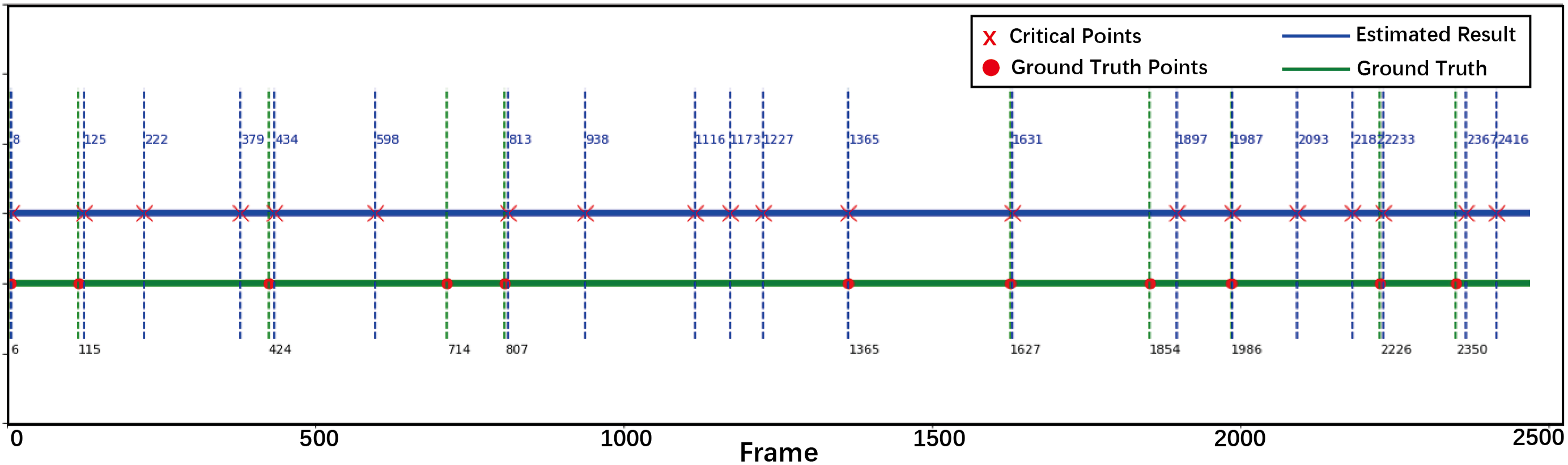}
  \caption{ An example of the segmentation result using vision and kinematics data. (Data Source: Knot Tying File I002, JIGSAWS)}
 \vspace{-0.4cm}
  \label{final knot}
\end{figure*}

\begin{figure*}[htpb]
  \centering
   \captionsetup{font=footnotesize,labelsep=period}
  \includegraphics[width = 0.9\linewidth]{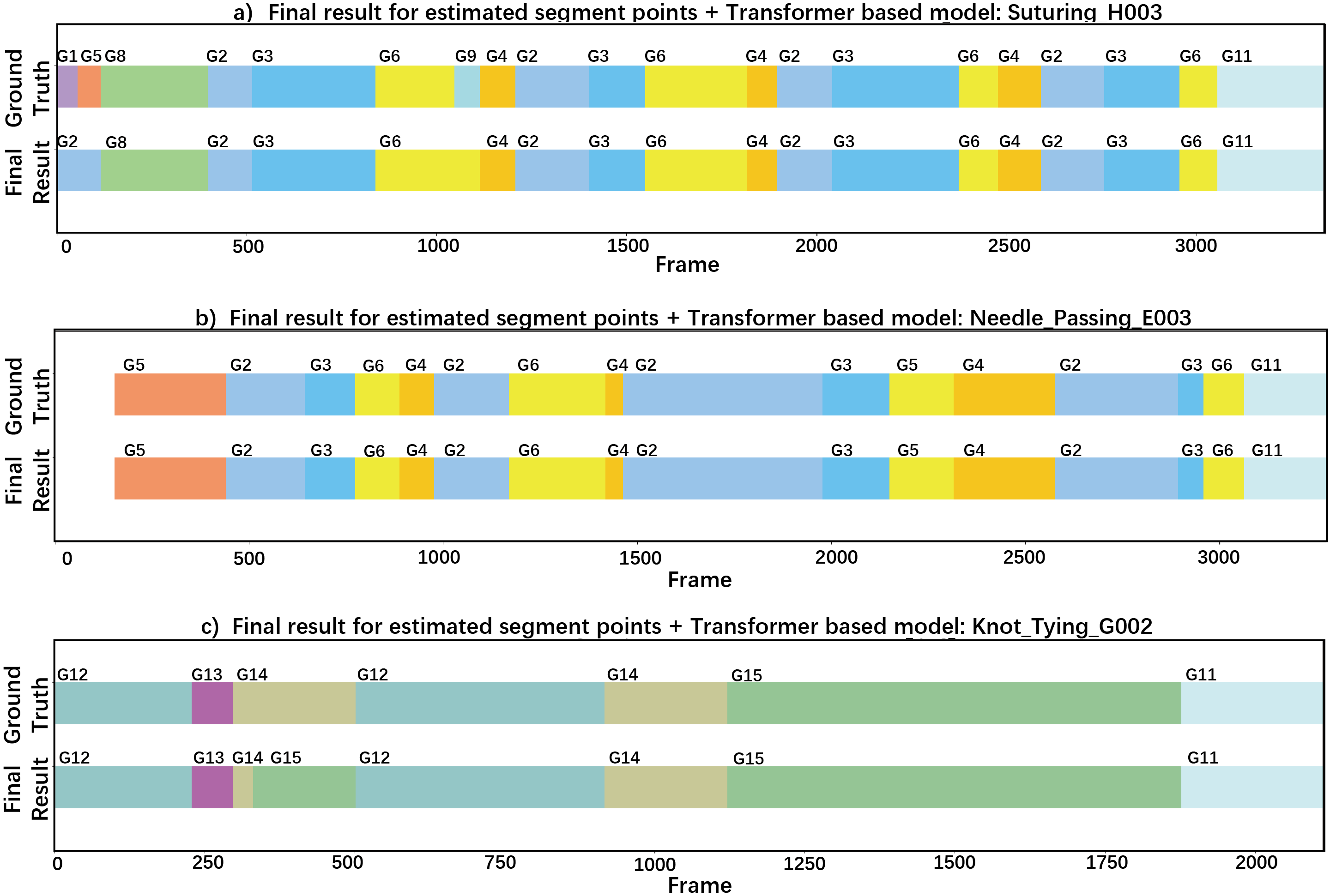}
  \caption{Examples of surgical gesture classification results.}
     \vspace{-0.3cm}
  \label{class K}
\end{figure*} 

\subsubsection{Final Classification Results}

The segmentation results are utilized to partition the entire video into segments, generating a new dataset that enables the classification of each segment for the surgical gesture recognition task. The results of this task are summarized in Table \ref{predict}.
Fig. \ref{class K} provides examples of classification results with estimated segmentation points.  This indicates that the Transformer-based model  can be used to refine the results generated by the unsupervised methods and improve the classification accuracy.

Based on the analysis of the confusion matrix (see supplementary materials provided by our website), we observed that Gesture 10 (Loosening more suture) can be easily identified. Conversely, Gesture 5 (Moving to center with  needle in grip) has the lowest classification accuracy, with 25\% of Gesture 5 instances being misidentified as either Gesture 1 (Reaching for the needle with the right hand) or Gesture 2 (Positioning the needle). A potential reason for this could be that the Transformer-based model can only accept 30 frames as synchronous input, and different gestures might involve similar motion within these 30 frames. Since Gestures 1 and 2 involve the state of `having the needle in grip' at the center of the image, they might be misclassified as Gesture 5.

\begin{table}[htpb]
\centering
\caption{Summary of Classification Results}
\begin{tabular}{ c  c | c  c  } 
\hline
\textbf{Metrics} & \textbf{Result} & \textbf{Metrics} & \textbf{Result} \\ \hline
Recall\_micro & 0.868 & Precision\_micro &  0.868 \\
\hline
Recall\_macro & 0.829 & Precision\_macro & 0.830 \\ 
 \hline
Recall\_weighted & 0.867 & Precision\_weighted & 0.879 \\ \hline
Accuracy & 0.856 & & \\ \hline
\end{tabular}
 \vspace{-0.4cm}
\label{predict}
\end{table}

\subsection{Comparisons with State-of-the-Art Methods}

\begin{table}[htbp]
\centering
\caption{Comparisons with other State-of-the-Art Approaches}
 \begin{tabular}{c | c | c } 
 \hline

\textbf{Method} & \textbf{Metrics} & \textbf{Scores} \\  \hline
Space and variance & Recall/Precision/F1 & 0.652/ 0.92/0.754 \\ \hline
Soft-UGS & Recall/Precision/F1 & 0.74/0.71/0.72 \\ \hline
LSTM &  Acc & 0.805 \\  \hline
TCN & Acc & 0.796 \\ \hline
Zero-shot & Acc & 0.56 \\ \hline
CNN+LC-Sc-CRF & Acc & 0.766 \\ \hline
\textbf{Ours} & Acc & \textbf{0.856} \\ \hline
\end{tabular}

\label{compare}
\end{table}

LSTM \cite{dipietro2016recognizing} and TCN \cite{lea2016temporal} are considered state-of-the-art supervised learning methods for classification. In addition, Bimanual space and variance \cite{tsai2019unsupervised}, Soft-UGS \cite{fard2016soft}, and Zero-shot \cite{kim2021daszl} represent state-of-the-art unsupervised learning techniques. 

In this paper, these methods serve as baselines for comparative studies and are all evaluated using the Suturing Task. The results presented in Table \ref{compare}, reveal that our proposed method outperforms the baseline techniques in terms of accuracy.

\section{Discussions and Future Work}

We propose a hierarchical semi-supervised learning framework for surgical gesture segmentation and recognition. In the first hierarchy, we identify the critical points from kinematics data and video data and then determine the potential segmentation points. In the second hierarchy, we use an unsupervised learning approach to cluster the potential segmentation points and determine the final segmentation points.
By leveraging implicit information from both kinematic and video data, the proposed method is capable of successfully identifying segmentation points and recognizing gesture labels. The effectiveness of our proposed method was verified using the JIGSAWS dataset, achieving an impressive accuracy of 0.856 for classification and an F1 score of 0.623 for segmentation.

In the future, we plan to expand the algorithm by incorporating self-supervised methods and leveraging sim-to-real learning techniques to further eliminate the need for labeling real operation data for surgical gesture recognition tasks \cite{zhang2022human}. Furthermore, we aim to apply the proposed method to support the development of automation in robotic surgery based on learning from demonstration.

\bibliographystyle{IEEEtran}
\bibliography{ref}

\end{document}